\pdfoutput=1
\documentclass[11pt]{article}
\usepackage{acl}

\usepackage{times}
\usepackage{latexsym}
\usepackage{amsmath}
\usepackage[T1]{fontenc}
\usepackage[utf8]{inputenc}

\usepackage{microtype}

\usepackage{inconsolata}
\usepackage{enumitem}
\usepackage{graphicx}
\usepackage{adjustbox}
\usepackage{tabularx}
\usepackage{booktabs}
\usepackage{array}
\usepackage{longtable}
\usepackage{subfigure}
\usepackage{algorithm}
\usepackage{algpseudocode}

\usepackage{array}   % for \newcolumntype macro
\newcolumntype{L}{>{$}c<{$}} % math-mode version of "c" column type
\usepackage{float}
\usepackage{tablefootnote}

\title{Aligning Sentence Simplification with ESL Learner's Proficiency for Language Acquisition}

\author{
 \textbf{Guanlin Li\textsuperscript{1,2}},
 \textbf{Yuki Arase\textsuperscript{2}},
 \textbf{Noël Crespi\textsuperscript{1}}\\
 \textsuperscript{1}Samovar, Telecom SudParis,
 Institut Polytechnique de Paris, France\\
 \textsuperscript{2}School of Computing, Institute of Science Tokyo, Japan\\
 \texttt{ \{guanlin\_li, noel.crespi\}@telecom-sudparis.eu, arase@c.titech.ac.jp}
}

\begin{document}
\maketitle
\begin{abstract}

Text simplification is crucial for improving accessibility and comprehension for English as a Second Language (ESL) learners. 
This study goes a step further and aims to facilitate ESL learners' language acquisition by simplification. 
Specifically, we propose simplifying complex sentences to appropriate levels for learners while also increasing vocabulary coverage of the target level in the simplifications. 
We achieve this without a parallel corpus by conducting reinforcement learning on a large language model. 
Our method employs token-level and sentence-level rewards, and iteratively trains the model on its self-generated outputs to guide the model to search for simplification hypotheses that satisfy the target attributes. 
Experiment results on CEFR-SP and TurkCorpus datasets show that the proposed method can effectively increase the frequency and diversity of vocabulary of the target level by more than $20\%$ compared to baseline models, while maintaining high simplification quality. 
\footnote{Codes available at \url{https://github.com/JumpyPizza/align-sentence-simplification-with-ESL-learner}}

\end{abstract}
\section{Introduction}
Controlled text simplification considers audience-targeted attributes when generating simplified texts, so that the generated texts do not only meet the criteria of simplicity, but also preserve desired attributes for the targeted audiences. 
Recent studies on controlled text simplification aimed to help reading comprehension for language learners and employed school grade levels annotated in the training corpus as the simplification target \cite{scarton-specia-2018-learning, sheang-saggion-2021-controllable,agrawal-carpuat-2023-controlling} or text features (number of words, character-level Levenshtein similarity etc.) between source and target sentences \cite{nishihara-etal-2019-controllable, martin-etal-2020-controllable}. 

Different from these studies, we aim to aid language learning and education for English as a Second Language (ESL) learners by simplifying sentences while preserving desirable attributes for language acquisition. 
We use the Common European Framework of Reference for Languages (CEFR), the world standard definition of language proficiency. 
Our method is motivated by two classic L2 learning theories: the input hypothesis \cite{krashen1981second} and frequency effect \cite{ellis2002frequency}. 
The input hypothesis stated that in order for the language acquisition to happen, the textual input which is either too simple or too complex for learner comprehension will not be useful for acquisition. If a learner’s current competence is $i$, then comprehensible input should contain both $i$ and $(i + 1)$ content \cite{mitchell2019second}. Frequency theory holds that the frequency of the words and phrases in the input is a key determinant of acquisition, and words with higher frequency in the usage tend to be easier to acquire \cite{ellis2009construction}. 
The key challenge here is the lack of a parallel corpus for training that should provide complex-simple sentence pairs labelled their levels. 
Parallel sentences of this kind are naturally scarce, and worse, annotation of difficulty levels, in particular, CEFR, is non-trivial and requires language education experts~\cite{arase-etal-2022-cefr}.

To achieve sentence simplification for aiding language learning without a parallel corpus, we propose reinforcement learning on a pre-trained large language model (LLM). 
% method to control the simplification outputs according to the learners’ language ability without the need for a parallel corpus. 
Based on the aforementioned L2 learning theories, the proposed method simplifies the complex sentences to the one corresponding to the learner's proficiency level or one level higher ($i$ and $i+1$ levels) and increases the coverage (frequency and diversity) of the corresponding level’s vocabulary in the generated simplifications. 
Specifically, we reformulate the controlled simplification task as a lookahead search problem: in the decoding step $t$, the model searches for the token that satisfies the target vocabulary constraint while also ensuring that future tokens 
%in $t+1$ to $t_{EOS}$ 
increase the target vocabulary coverage as much as possible, and the final hypothesis falls into the desired CEFR level. 
We combine a simple word-match-based heuristic with the supervised sentence-level signal to guide decoding and train the model iteratively using gradient policy optimization to memorize the search strategy that maximizes the overall reward. 
Remarkably, we eliminate the need for a parallel corpus by utilizing LLMs' language generation capacity for simplification via reinforcement learning. 
% In this process, only target-audience-related attributes are required and the need for complex-simple parallel sentences with CEFR annotation is eliminated. 
Experimental results show that the method significantly increases the coverage and diversity of the target vocabulary in the outputs by up to $20\%$ compared to the baselines, while maintaining high simplification quality.

Our primary contributions are twofold. 
First, we propose the sentence simplification method that aligns generated simplifications with ESL learners’ proficiency level on word, phrase and sentence levels and preserves attributes effective for facilitating language learning. 
Second, our method is easy to deploy and does not require a parallel corpus that is often expensive to create.

\section{Related Work}
We briefly summarize two lines of simplification methods, controlled simplification and reinforcement learning based simplification. %, in the following part.

% \textbf{Control token/Soft prompting} 
\paragraph{Controlled Simplification} attaches tokens or prompts to the input %to represent attributes. Consequently, the tokens can be used 
to control the simplification-related attributes during generation \cite{yang-etal-2023-tailor,agrawal-carpuat-2023-controlling, sheang-saggion-2021-controllable,martin-etal-2020-controllable, martin-etal-2022-muss, scarton-specia-2018-learning, chi-etal-2023-learning}. 
While these methods learn to control levels of simplified sentences using a parallel corpus with annotated difficulties, our method controls attributes useful for language learning without a parallel corpus. 
As opposed to the training-time controlling, \citet{kew-ebling-2022-target} adopted FUDGE \cite{yang-klein-2021-fudge}, which adjusts the logits of the text generation model during decoding using a classifier, to directly control the attribute of the simplification in the decoding time.

\paragraph{Reinforcement Learning based Simplification} has explored
% unsupervised 
controllability by defining rewards based on simplicity-related criteria %and dynamically determining the trade-offs for these objectives during generation using reinforcement learning (RL) 
\cite{zhang-lapata-2017-sentence,guo-etal-2018-dynamic,nakamachi-etal-2020-text,laban-etal-2021-keep}. The rewards for the objectives are constructed using supervised or unsupervised evaluation metrics for simplicity, adequacy and fluency. 
In contrast, we aim to control attributes useful for language learning and education. 
Furthermore, while RL tends to suffer from unstable training and sensitivity to the choice of hyperparameters, our method achieves training stability by adopting entropy regularization in the model optimization process and introducing a dynamic reward that adjusts based on the data distribution.

\section{Problem Definition}\label{sec:method}
% We aim to facilitate language learning by generating text simplifications targeted at English learners with constrained knowledge of the language. In reality, different individuals have varied levels of knowledge for the language. 

We aim to facilitate language learning by simplification targeted at ESL learners. 
% In reality, different individuals have varied levels of knowledge for the language,
In this study, we use CEFR levels as a representative measure for the learners' proficiency and model the target level based on the vocabulary\footnote{https://www.englishprofile.org/wordlists/evp}  (words, phrases, idioms) 
and sentence 
% usage (as indicated in CEFR\footnote{https://www.coe.int/en/web/common-european-framework-reference-languages/table-1-cefr-3.3-common-reference-levels-global-scale}). 
CEFR levels\footnote{https://www.coe.int/en/web/common-european-framework-reference-languages/table-1-cefr-3.3-common-reference-levels-global-scale}.

Our problem is thus defined as follows. 
We assume that learners know their own CEFR level $i$. 
Given a sentence above the learner's level $i$, we generate its simplified version that (a) contains as much vocabulary of the level $i$ and $i+1$ as possible, and (b) corresponds to the target (learner's) level $i$ at the sentence level.  %matches the level of the generated simplifications to the target levels. 
% based on the judgment determined by expert English teachers. 

\subsection {Constraint Formalization} \label{sec:task}
Generating simplified texts subject to vocabulary constraints can be approached as a lexical-constrained text generation task \cite{zetsu-etal-2022-lexically}. 
Traditionally, lexical constraints in text generation involve a \emph{short} list of required words, which \citet{lu-etal-2021-neurologic} expressed as a Conjunctive Normal Form (CNF), such as $\underbrace{(D_1 \lor D_2 \lor \cdots )}_{C_1} \land \cdots \land \underbrace{(D_{m-1}  \lor D_m)}_{C_m}$ in which $D_m$ stands for a single constraint, and all clauses must be satisfied, imposing hard constraints on the generation process.

In our setting, however, this formulation is no longer applicable because the vocabulary constraint is \emph{as large as the size of the vocabulary of a specific level}. 
In addition, we aim to satisfy \emph{as many clauses as possible}. 
Therefore, we formalize constraints as Disjunctive Normal Form (DNF), indicating words and phrases suitable for the target proficiency level: $D = \underbrace{(D_1)}_{C_1} \lor \underbrace{(D_2 \land D_3 \land \cdots )}_{C_2} \lor \cdots \lor \underbrace{(D_m)}_{C_m}$, where the form stands for the word list of the target language level, a single $D_m$ represents word and the conjunctive clauses represent several words, namely phrases. 
Notably, this form of constraints allows for the control of discontinuous phrases, which is difficult in previous methods. 

\subsection{Optimization Function}
Based on the DNF constraints, our task imposes soft constraints that aim to include as many clauses as possible. 
% For a collection of complex sentences, 
Given the simplification hypotheses $\{ \text{seq}_1, \text{seq}_2, \ldots,\text{ seq}_n \}$, the goal is to maximize:
\begin{equation}\label{eq:goal}
    \sum_{j=1}^{m} \sum_{k=1}^{n} \text{count}(C_j, \text{seq}_k),
\end{equation}
where $\text{count}(C_j,  \text{seq}_k)$ indicates the number of clauses $C_j$ satisfied by $ \text{seq}_k$. 
Consequently, the target during the generation process is to search for the next token that: 

\begin{itemize}[noitemsep,topsep=1pt,parsep=0pt,partopsep=0pt,]
 
    \item simplifies the original text;
    \item is contained in $\exists C_i \in D$; 
    \item leads to future tokens that satisfy $C_i$; 
    \item leads to complete phrases or phrases with slots (discontinuity) that satisfy $C_i$.
\end{itemize}

\section{Proposed Method}
\begin{figure*}[t!]
    \centering
    \includegraphics[width=\textwidth]{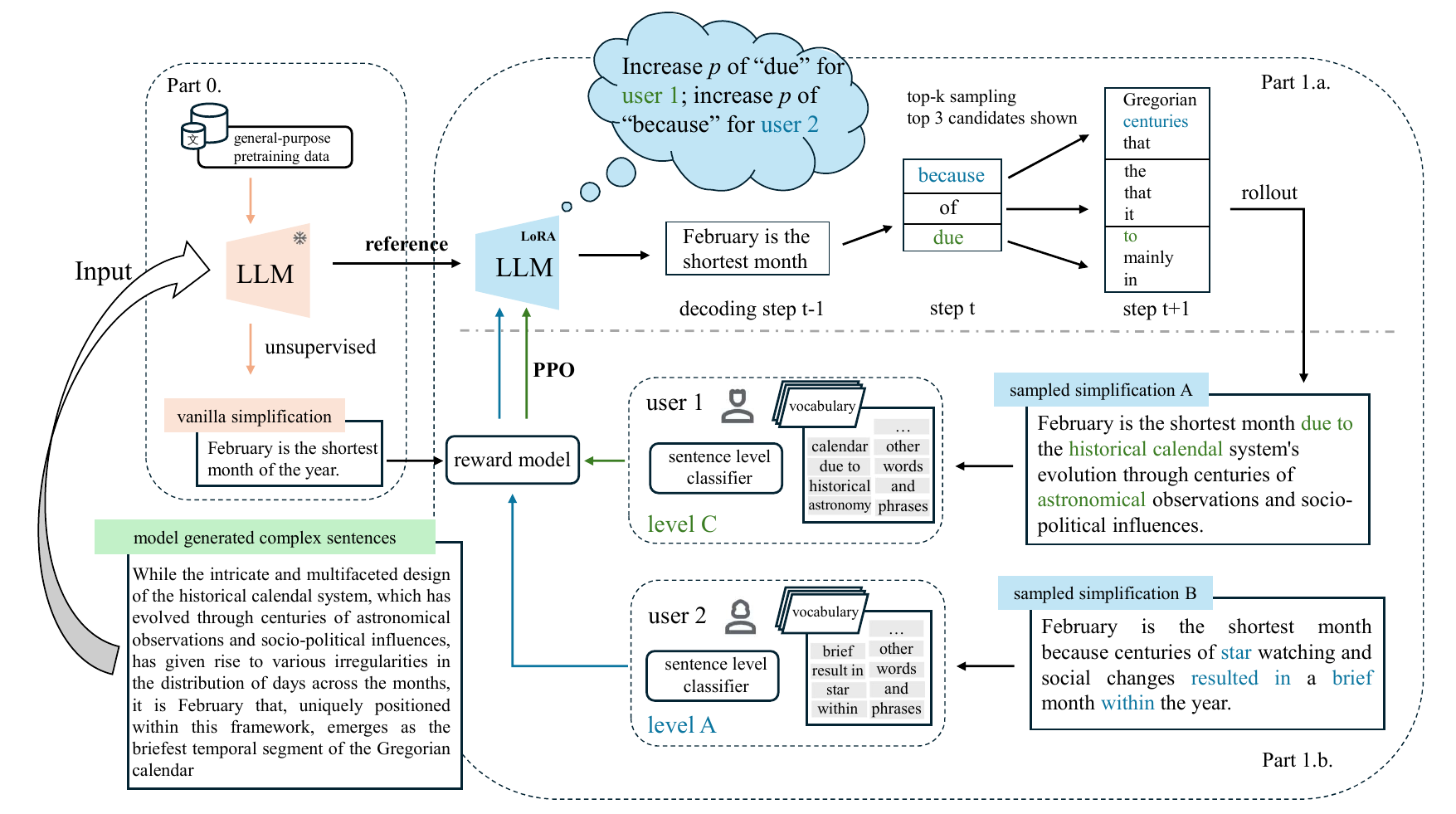}
    \caption{(better viewed in color) The overall framework of the proposed method: the simplification model is initialized from a pretrained large language model which is also used as a frozen (\includegraphics[height=\fontcharht\font`X]{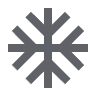}) reference model to provide entropy regularization (part 0.); top-k sampling is adopted in the decoding process to sample varied simplifications for the complex sentence (part 1.a.); the generated simplifications are evaluated based on the language proficiency level (vocabulary level and sentence level) of the target audience, which is used as rewards to update the simplification model (part 1.b.) to adopt better decoding strategy. % to generate simplifications that suit the target users. 
    }
    \label{fig:framework}
\end{figure*}

To search for a hypothesis that better satisfies predetermined constraints, some previous methods use rollout in decoding that generates partial future sequences \cite{chaffin-etal-2022-ppl, lu-etal-2022-neurologic}. 
% This may work for smaller models, but they become infeasible for large models. 
These methods become infeasible for large models due to the inefficiency of sampling in decoding time and handling the large vocabulary constraints in our task. %since they apply the lexical constraints in CNF which only considers a small list of lexical terms. 
To effectively and efficiently search for the tokens that satisfy our constraints, we instead consider sampling in the training time and formulate the lookahead search problem using RL search~\cite{fickinger2021scalable} (see Fig.~\ref{fig:framework}). 

\subsection{RL Search}
Consider the text generation process as a Markov Decision Process (MDP), at each timestep $t$ of the generation, the language model observes a state $s_t$, which is the generated partial sequence $ \text{seq}_{t-1}$, and takes an action $a_t$ to choose the token from its vocabulary. When the EOS token is reached, a reward $R$ for the generated sequence is calculated and used to update the model. %given by the heuristic to measure the number of clauses in DNF that the sequence satisfies. 
In this setting, the language model is the policy function that 
searches for a token $v_i \in \mathcal{V}$ where $\mathcal{V}$ is the vocabulary,
% that gives the distribution $\pi(a|s)$ over actions $a$, 
and we can use any policy gradient algorithm to guide the language model to search for the generations that maximize the constraint satisfaction. 
Algorithm~\ref{alg:training} indicates our training procedure. 
% Combining the sentence level reward and the target vocabulary constraints, the process of the proposed method is then as follows:
\begin{algorithm}[t!]
\caption{Training Procedure}
\label{alg:training}
\begin{algorithmic}[1]
% [noitemsep,topsep=1pt,parsep=0pt,partopsep=0pt,]
    \Require Complex sentences;
    \State Generate simplified texts from the complex sentences using current policy (rollout);
     \State Evaluate the current policy and produce rewards to guide the search;
    \State Optimize the policy model using the rewards
    \State Iteratively perform steps 1-3 till converge.
\end{algorithmic}
\end{algorithm}

\subsection{Policy Model}\label{sec:policy_model}
The policy model generates a simplified sentence $\text{seq}$ given a complex counterpart as a prompt $\text{pmt}$. 
The policy model is initialized from an instruction tuned language model, which unsupervisedly provides robust text simplifications \cite{kew-etal-2023-bless}.

By design, the rewards for the policy model across different proficiency levels are varied. For instance, given the same model response, a positive reward for C level could correspond to a negative reward for A level. Therefore, using the original language model as the backbone, we train separate copies of the policy model for A, B and C levels by adding and updating distinct LoRA parameters to the backbone parameters \cite{hu2022lora}, while keeping the backbone frozen.

\subsection{Reward Models}
Inspired by the L2 learning theories, we design two types of rewards at lexical and sentence levels.

 \subsubsection{Lexical Constraint Reward}
We use a simple heuristic to guide the search for generations that satisfy the lexical constraints:
\begin{equation} \label{eq:cnt}
    H(\text{seq}) = \sum_{C_j \in D} r (\text{count}(C_j, \text{seq})),
\end{equation}
where $C$ is a clause from  $D$, 
% the DNF, 
$r$ denotes the reward according to the number of satisfactions of $C$ in $\text{seq}$, and $H$ denotes the reward score for the generated sequence $\text{seq}$ in the current decoding step.
To calculate the match counts, we remove basic stop words from the sentence after lemmatization.

% For each count, the reward $r_H$ is set to be 1. 
As a simple baseline, we define $r$ as a constant value $1$ for word and $1.5$ for phrase
% Additionally, $r$ is set to be a higher constant value when C is a phrase, 
to encourage the model to generate more phrases and idioms. 
However, we found that this simple baseline is easily hacked by the model after a few steps of training, i.e., the model only generates a limited set of frequent words that were learnt to produce rewards.
To encourage the model to explore more diverse words and improve the overall coverage of target-level words in the generations, the reward should intuitively encourage maximizing the entropy for the clauses in $D$, 
so that all the clauses are evenly distributed. Accordingly, we adjust the reward $r$ for the count of $C_j$ as a \textbf{dynamic reward}:
\begin{equation}\label{reward}
     r = \begin{cases} 1 & \text{if } 0 \le p_j < \frac{1}{m}, \\ e^{-\alpha p_j} & \text{if } \frac{1}{m} \le p_j \le 1, \end{cases} 
\end{equation}
where $p_j$ is:
\begin{equation}\label{cnt_c}
    p_j = \frac{\sum_{k=1}^{n} \text{count}(C_j, \text{seq}_k)}{\sum_{k=1}^{n} \sum_{j=1}^{m} \text{count}(C_j, \text{seq}_k)},
\end{equation}
to discourage the model from exploiting the same clause. 
Here, $m$ denotes the total number of clauses in $D$ and $\alpha$ is a constant to adjust the penalty degree for too frequent clauses. 
Eq. \ref{cnt_c} is calculated after each epoch and the reward is adjusted accordingly.
If matched clauses are above the target level, we give a constant negative score of $-1$.

\subsubsection{Sentence Level Reward}    
To go beyond words and guide the simplification model’s search for a sentence of the target level, we incorporate a sentence-level reward model by simulating human experts’ judgment for the sentence’s CEFR level. 
We use pairwise ranking loss to train the reward model,
since the class distribution for the CEFR-SP data is imbalanced \cite{arase-etal-2022-cefr}. %, training a classification model using CE would 
% amplify the problem. 
%bias the model towards the majority class.
The ranking loss has been shown to be able to encourage the model to only pay attention to the focused class \cite{henning-etal-2023-survey}, thus may mitigate the class imbalance problem. 

Consequently, we construct sentence pairs prioritizing the level we focus on generating: for a collection of sentences $\mathcal{S} = \{s_1, s_2, \ldots, s_n\}$, each sentence $s_i$ is evaluated by human experts and annotated with a language level $l$. Given the level we want to generate, we select the sentences with the target level $\mathcal{S}_{\text{tgt}} = \{ s_i \in \mathcal{S} \mid l_i = \text{level}_{\text{tgt}} \}$, and randomly sample sentences from other levels to construct a negative set $\mathcal{S}_{\text{non-tgt}} = \{ s_j \in \mathcal{S} \mid l_j \neq \text{level}_{\text{tgt}} \}$. Then, we construct sentence pairs $\mathcal{P} = \{ (s_i, s_j) \mid s_i \in \mathcal{S}_{\text{tgt}}, s_j \in \mathcal{S}_{\text{non-tgt}} \}$ by randomly selecting from $\mathcal{S}_{\text{tgt}}$ and $\mathcal{S}_{\text{non-tgt}}$. 

Notably, we do not require the pair to be parallel; they just need to be at different levels. %, that is, to be different simplified versions of the same sentence. 
By this design, we disentangle the adequacy requirement for the simplification from the target-level search process. 
The former is handled by the underlying LLM, and the latter is dealt with by the reward model by level judgment.

With the constructed sentence pairs, we train a sentence-level reward model $r_\theta$. The training objective is to minimize loss: 
\begin{equation}
    \mathcal{L}(\theta) = - \sum_{(s_i, s_j) \in \mathcal{P}} \log \sigma(r_\theta(s_i) - r_\theta(s_j))
\end{equation}
where $\sigma$ is the sigmoid function. 
After training the reward model, for a generated sentence $\text{seq}$, we take $r_l = \sigma(r_\theta(s))$ as the reward, and use a linear combination of the lexical reward and the sentence-level reward as the overall reward: 
\begin{equation}
    R = \lambda r + \gamma r_l
\end{equation}

% $r = ar+br_l$. 

\subsection{Stabilized RL Training}
The original instruct-tuned model is used as a frozen reference model, providing an entropy regularization for the updated policy model to ensure training stability during the search process. 
Specifically, the simplification $\text{seq}^\prime$ produced by the 
% vanilla model 
frozen backbone model
$f^\prime$ is added as an entropy regularization to the overall reward:
\begin{equation}
    R^\prime = R-\log(p_{(f(\text{seq}|pmt)}/p_{(f^\prime(\text{seq}^\prime|pmt))}).
\end{equation}
By doing so, we may keep the LLM's strong paraphrasing ability while letting it acquire controllability in CEFR levels.   

The policy model $f$, namely the simplification model is then updated to search for the generations that maximize the reward. In this study, we adopt Proximal Policy Optimization \cite{schulman2017proximal} to update the policy model, which achieves stable training and faster convergence.

\section{Experiment Settings}
We aim to evaluate the effectiveness of the proposed model in generating high-quality simplifications that align with the target vocabulary and sentence-based CEFR level. 
This section provides details of the experiment settings.
%We also provide descriptions of the resources used for training and evaluation, the implementation details, and the evaluation metrics.

\subsection {Resource and Implemetation}\label{sec:resource}

 \begin{table}[t!] \centering 
 \begin{adjustbox}{width=0.9\linewidth}
\begin{tabular}{lrrrrrr} 
\toprule & A1 & A2 & B1 & B2 & C1 & C2 \\ 
\midrule Train & 248 & 1284 & 2479 & 2226 & 889 & 52 \\
Val & 79 & 276 & 485 & 336 & 149 & 40 \\ 
Test & 71 & 289 & 540 & 369 & 150 & 39 \\ 
\bottomrule 
\end{tabular} 
\end{adjustbox}
\caption{Statistics on CEFR-SP w/o Newsela} 
\label{tab:cefrsp}
 \end{table}

 \begin{table*}[t!]
    \small
    \centering
   \begin{adjustbox}{width=0.9\textwidth}
    \begin{tabular}{l L L L L L L L} \hline
     \textbf{CEFR-SP} &  \text{A-Frequency} & \text{A-Diversity} & \text{B-Frequency}  & \text{B-Diversity} & \text{C-Frequency}  & \text{C-Diversity}     \\ \hline
     Reference & 0.292 & 0.527 & 0.283 & 0.465  & 0.080 & 0.102  \\ 
      phi3-3b-vanilla &0.252 & 0.665 & 0.215 &0.435  &0.041 & 0.172  \\ \hline
       T5+grade-A & 0.194 & 0.438 & 0.269 & 0.271  &0.072 & 0.114  \\
       FUDGE-A & 0.257 & 0.215 & 0.207 & 0.069  &0.043 & 0.018 \\
       \textbf{phi3-A} & \textbf{0.299}  & \textbf{0.684} & 0.196 & 0.403  & 0.038 & 0.141 \\ \hline
       
       T5+grade-B & 0.204 & 0.447 & \textbf{0.275} & 0.266  &0.069 & 0.110  \\
       FUDGE-B & 0.223 & 0.226 & 0.231 & 0.084  &0.049 & 0.027 \\
        \textbf{phi3-B} & 0.151 & 0.677 & 0.262 & \textbf{0.538}  & 0.064 & 0.251 \\ \hline
        
      T5+grade-C & 0.203 & 0.441 &  0.276  &0.271 & 0.074 & 0.114 \\
       FUDGE-C  & 0.239 & 0.217 & 0.220 & 0.077  &0.052 & 0.025 \\
        \textbf{phi3-C} & 0.171 & 0.658 & 0.263 & 0.275  & \textbf{0.189}  & \textbf{0.365} \\ 
         \hline
    \end{tabular}
   \end{adjustbox}
    \caption{Results on target attribute controllability on CEFR-SP-Test. For ``Reference'', the frequency and diversity metrics were calculated using a subset of each grade level to show distributions in sentences of specific levels. }
    \label{tab:cefr1}
\end{table*}

\paragraph{Sentence CEFR Level} To train the sentence-level reward model,
we used CEFR-SP \cite{arase-etal-2022-cefr}, which provides labels of six CEFR levels for a total of $17k$ sentences annotated by experts.  
We only used the publicly available subset from the dataset (excluding data based on Newsela \cite{xu-etal-2015-problems}), which resulted in $10k$ sentences with labels.
The statistics of the dataset are described in Table\ref{tab:cefrsp}. %Appendix \ref{sec:appendix-training}. 
We fine-tuned the  GPT-2 \cite{radford2019language} using the annotated CEFR levels.

\paragraph{Vocabulary List}
For the lexical constraint reward model, we need vocabulary lists per CEFR level. 
We downloaded the English Vocabulary Profile (EVP) data\footnote{https://www.englishprofile.org/wordlists/evp} and used it as a dictionary of words and phrases annotated with their corresponding CEFR levels\footnote{EVP assigns CEFR levels to each word sense, so the same word can appear at different levels depending on its meaning. For simplicity, we did not conduct word sense disambiguation and assigned the lowest level.}.
% To calculate the match counts for lexical constraint reward, we remove basic stop words from the sentence, and lemmatize both words in the sentence and in the list. 
Since our goal is to generate the simplifications in $i$ and $i+1$ levels, we always aggregate the vocabulary lists in two levels. For clarity, we consider A1$+$A2, B1$+$B2, and C1$+$C2 levels. In total, we got $1076$ words for A level, $3823$ words for B level, $3612$ words for C level. 

\paragraph{Complex Sentence Collection}
We trained the policy model to iteratively learn to search for a hypothesis that maximizes rewards based on its own generations. 
The only requirement for our training corpus is a supply of complex sentences that warrant simplification, because sufficiently simple sentences without the need for simplification may disturb the learning. 
\citet{cegin-etal-2023-chatgpt} showed that large language models are highly capable of paraphrasing. 
Following this study, we used GPT-4\footnote{https://openai.com/index/gpt-4/} to synthesize complex sentences from the CEFR-SP training set to create our training corpus. 
We manually prepared prompts to ensure that the outputs are always at least as complex as the highest C2 level. 
More details are in Appendix \ref{sec:appendix-complex}.

We trained separate models for A, B and C levels since different levels require different rewards (see Section~\ref{sec:policy_model}). 
For computational efficiency, we adopted a relatively small Phi-3-mini-3b model \cite{abdin2024phi}. 
% The complex sentences are generated from the CEFR-SP dataset using GPT-4 \footnote{https://openai.com/index/gpt-4/}.
% Before training, a GPT-2 model \cite{radford2019language} was trained on the CEFR-SP sentence labels as the sentence level reward model; during training, only the complex sentences are used as inputs.  
More implementation details can be found in Appendix \ref{sec:appendix-training}.

\subsection{Evaluation Datasets}
To evaluate the simplification outputs, we need parallel corpora of complex and reference simple paraphrases. Below describes the resources we used for the evaluation.

\paragraph{CEFR-SP-Test}
As the formal evaluation dataset, we used CEFR-SP. 
We expanded its test set to be parallel because CEFR-SP is a non-parallel corpus. 
Specifically, we generated complex sentences for each sentence in the CEFR-SP test set using the same method described in Section~\ref{sec:resource}. 
These complex sentences were input to models, and outputs were evaluated by comparing them to the original CEFR-SP sentences as references.

\paragraph{TurkCorpus} To assess the applicability of the proposed method for a general simplification task, we also evaluated models on another widely-used dataset, TurkCorpus \cite{xu-etal-2016-optimizing}. We used the test set of the corpus, including $359$ complex sentences, each has $8$ human-written simplified sentences as references. 
It should be noted that TurkCorpus does not provide any level annotations. 
% The evaluation on the TurkCorpus is zero-shot: the models trained on CEFR-SP complex sentences are directly evaluated using TurkCorpus test set.

\subsection{Evaluation Metrics}
We evaluated simplification outputs from two perspectives: \textbf{simplification quality} and \textbf{target audience attributes} by both automatic and human assessments. 
Simplification quality was assessed across three dimensions: \textbf{Simplicity; Fluency; and Adequacy}.         
As automatic metrics for simplicity, we employed LENS \cite{maddela-etal-2023-lens} and SALSA \cite{heineman-etal-2023-dancing}, which are two recently proposed model-based evaluation methods. 
For fluency and adequacy, we employed an instruction-tuned language model as an off-the-shelf evaluation model, which was shown to be effective in automatic translation quality evaluations \cite{kocmi-federmann-2023-large}. 
Target audience attributes were measured in terms of \textbf{target vocabulary coverage} and \textbf{sentence CEFR level},
% Vocabulary coverage includes both frequency and diversity of target vocabulary. 
in which vocabulary coverage includes both frequency and diversity of target vocabulary. 
% For determining sentence CEFR-level, we reused our reward model. 
For the evaluation of sentence CEFR-level, we used human evaluation. 
% For sentence level, we measure the absolute level using a classification model trained on the CEFR-SP dataset, and report the binary accuracy as the sentence level score. 
% Nonetheless, human evaluation is crucial to assess the quality of simplification. 
For more details on evaluation metrics, please refer to Appendix \ref{sec:appendix-metrics}.

\subsection{Baselines}
Overall, we choose two lines of work as the baselines for comparison.  

\paragraph{Controlled Simplification}
There are limited variants in controlled simplification methods which mostly employ control tokens with supervised learning. 
Based on previous literature, we implemented two baselines for controlling the target level of the simplified texts: a supervised baseline of T5+grade \cite{scarton-specia-2018-learning} that attaches CEFR levels as control tokens and an unsupervised baseline of FUDGE that uses a discriminator at decoding time \cite{yang-klein-2021-fudge}.

% \textbf{Direct simplification}
\paragraph{Non-controlled Simplification}
The Turk corpus was used to evaluate the effectiveness of the proposed method in general simplification. 
As opposed to controlled simplification, this task does not consider controlling attributes, such as grade levels, during the simplification.
% For this line of models, we choose various methods presented in the EASSE package \cite{alva-manchego-etal-2019-easse} to compare with the proposed method on TurkCorpus to demonstrate the method's robustness.
For this line of models, we choose the following methods: DRESS \cite{zhang-lapata-2017-sentence}, DMASS \cite{zhao-etal-2018-integrating}, EditNTS \cite{dong-etal-2019-editnts}, ACCESS \cite{martin-etal-2020-controllable}, IterativEdit \cite{kumar-etal-2020-iterative}. 
We used outputs of these models shared in the EASSE package \cite{alva-manchego-etal-2019-easse}.   
In addition, we also compare the vanilla phi3-3b instruction-tuned model as a baseline, under zero-shot setting without fine-tuning on simplification.

\begin{table*}[t!]
    \small
    \centering
   \begin{adjustbox}{width=0.9\textwidth}
    \begin{tabular}{l L L L L L L L} \hline
     \textbf{TURK} & \text{A-Frequency} & \text{A-Diversity} & \text{B-Frequency}  & \text{B-Diversity} & \text{C-Frequency}  & \text{C-Diversity}    \\ \hline
     Reference & 0.176 & 0.229 & 0.227 & 0.132  & 0.056 & 0.046  \\ 
     phi3-3b-vanilla &0.166 & 0.180 & 0.177 &0.083  &0.034 & 0.023 \\\hline
       T5+grade-A & 0.187 & 0.180 & 0.217 & 0.088  &0.051 & 0.028 \\
         
        FUDGE-A & 0.175 & 0.177 & 0.175 & 0.069  &0.034 & 0.018 \\
       
       \textbf{phi3-A} & \textbf{0.216} & \textbf{0.208} & 0.153 & 0.063  & 0.031 & 0.018 \\ \hline
       
       T5+grade-B & 0.201 & 0.190  & 0.217  &0.085 & 0.052 & 0.028 \\
    
    FUDGE-B & 0.163 & 0.177 & 0.178 & 0.077  &0.039 & 0.022 \\
       
        \textbf{phi3-B} & 0.126 & 0.201 & \textbf{0.330} & \textbf{0.112}  & 0.066 & 0.035 \\ \hline
        
      T5+grade-C & 0.187 & 0.194  & 0.225  &0.090 & 0.050 & 0.026 \\
       
       FUDGE-C  & 0.171 & 0.174 & 0.181 & 0.076  &0.037 & 0.019 \\        
       
        \textbf{phi3-C} & 0.151 & 0.178 & 0.193 & 0.092  & \textbf{0.091}  & \textbf{0.041} \\  \hline
        
    \end{tabular}
   \end{adjustbox}
    \caption{Results on target attribute controllability on TurkCorpus }
    \label{tab:turk1}
\end{table*}

\begin{table*}[t!]
    \centering
    \begin{tabular}{p{0.97\textwidth}}\hline
    \textbf{Complex Sentence}~~The considerable distance, compounded by Jamie's current condition of pregnancy, which inexorably engenders a state of increased fatigue, renders the prospect of ambulation to said location prohibitively challenging for her.\\\hline
\textbf{Ref. (level B)}~~It is too far for Jamie to walk to, especially because she is pregnant and easily exhausted.\\\hline
\textbf{Simplifications}  \\
Level A: Jamie is \textcolor{blue}{\textit{too}} \textcolor{blue}{\textit{tired}} \textcolor{blue}{\textit{to}} walk far because she is pregnant.\\

Level B: Jamie's pregnancy makes it very hard for her to walk to the location \textcolor{blue}{\textit{due to}} the long distance.\\

Level C: Jamie's pregnancy leads to \textcolor{blue}{\textit{fatigue}}, making it hard for her to walk to the distant place.\\\hline
    \end{tabular}
    \caption{A randomly selected example from the simplification result of the proposed method. The target vocabulary of the corresponding level is marked in \textcolor{blue}{\textit{italic}} font.}
     \label{tab:result demo}
\end{table*}

\section{Experiment Results}
This section analyses experiment results of automatic and human evaluations, and ablation study.  
\subsection{Automatic Evaluation Results}

% \subsection{Result Analyses}
\paragraph{Target Attributes }
Tables \ref{tab:cefr1} and \ref{tab:turk1} show the evaluation results for the target vocabulary coverage. 
These results demonstrate that 
compared to the baseline models, the proposed model significantly increases the frequency of target vocabulary in simplified sentences while also improving vocabulary diversity. 
Notably, the proposed method successfully increases the frequency and diversity of A and C-level vocabulary, which should be harder than B-level due to the scarcity of level A and C samples~\cite{arase-etal-2022-cefr}. 

\paragraph{Simplification Quality }
Tables \ref{tab:cefr2} and \ref{tab:turk2} show the evaluation results for the simplification quality. 
Overall, these results indicate that our models can produce high-quality simplifications, greatly outperforming the baseline models. 
Remind that our model does not have reward to encourage the model to follow the adequacy requirement.
% We attribute this to the benefits of using KL constraints and updating the policy model under the reference model's constraints.
We attribute these improvements to the benefits of using entropy regularization imposed by the reference model that allows the preservation of the high paraphrasing capability of LLMs.
% which ensure that the generation model do not deviate too much from a baseline generation.  
Table~\ref{tab:result demo} shows a randomly picked example of simplification by our method; Appendix~\ref{sec:appendix-results} provides more.

\subsection{Human Evaluation Results} \label{sec:human}
We perform a human evaluation to assess the simplification quality from human perspectives. 
We recruited three graduate-level students majoring in linguistics to perform the evaluation. 
The evaluators were first trained with the background knowledge and then given a guideline to evaluate the following aspects of the samples: fluency, simplicity, adequacy, and CEFR sentence level. 

We asked annotators to make binary judgements for fluency, simplicity, and adequacy. 
For sentence level, because CEFR-level judgements require expertise in language education, we simplified the task to collect reliable decisions. % to assess the model's ability to control simplification among different CEFR levels, 
We asked the evaluators to judge if a simplified sentence matches the desired sentence level (denoted as ``Level''). 
We showed a reference with its CEFR level and requested the evaluators to judge if the model output matches the reference's simplicity. 
% The results are shown in the "Level" column.
In addition, we asked them if a simplification output is preferable in terms of its CEFR level compared to the one generated by a model targeting a different level (denoted as ``Prefer''). 
For example, an evaluator judges if an output of the A-level model is preferable to that of the C-level compared to the A-level reference.\footnote{Preference score for the reference was judged by comparison with a sentence randomly chosen from another level} 
%, supposing that the task is an A-level simplification.
%newly added description here%%%%%%
% The results are shown in the "Prefer" column.
%%%%%%%%%%%
For each CEFR level, $30$ simplifications of the CEFR-SP-Test were randomly sampled and annotated ``Level'' and ``prefer'' judgements. 
We report the ratios of positive judgements as evaluation scores. 
% Evaluators were asked to choose ``yes'' or ``no" for each evaluation, and the resulting binary accuracy for the baseline and proposed models are presented in Table \ref{tab:human}. 
The details of the annotation guideline and interface are presented in Appendix \ref{sec:appendix-humaneval}. 

% \begin{table}[h]
%     \small
%     \centering
%    \begin{adjustbox}{width=1\linewidth}
%     \begin{tabular}{l c c c c c} \hline
%     Model & Simpicity  & Adequacy  & Fluency  & Prefer & Level \\ \hline
%      phi3-A & 1.0 &  0.76 & 0.90 & 0.67 &0.83 \\ 
%        phi3-B & 1.0 & 0.83 & 0.90 &0.70 &0.63 \\
%        phi3-C & 0.96 & 0.80 & 1.0 &0.80 &0.6 \\ \hline
%     \end{tabular}
%    \end{adjustbox}
%     \caption{Average human evaluation results.  Level denotes the match of level between reference sentence.}
%     \label{tab:human}
% \end{table}

\begin{table}[t!]
    \small
    \centering
   \begin{adjustbox}{width=1\linewidth}
    \begin{tabular}{l L L L L} \hline
    \textbf{CEFR-SP} & \text{LENS}  &  \text{SALSA}  &  \text{Fluency}  &  \text{Adequacy}    \\ \hline
     Reference & 43.57 &  59.54 & 0.829 & 0.624 \\ 
      phi3-3b-vanilla  & 63.37 & 74.18 & 0.897 & 0.538 \\ \hline
      
       T5+grade-A & 41.37 & 58.98 & 0.547 &0.291 \\
       
    FUDGE-A & 60.84 & 70.16 & 0.780 & 0.447 \\
    
    \textbf{phi3-A} & \textbf{67.29} & \textbf{76.23} & \textbf{0.827} & \textbf{0.604} \\ \hline
    
       T5+grade-B & 40.15 & 58.43 & 0.535 &0.290 \\
    FUDGE-B & 53.33 & 68.69 & 0.823 & 0.540 \\
    \textbf{phi3-B} & \textbf{64.61} & \textbf{72.21} & \textbf{0.871} & \textbf{0.768} \\ \hline

      T5+grade-C & 41.67 & 59.12 & 0.538 &0.277 \\
      
       FUDGE-C & \textbf{60.50} & 70.48 & 0.830 & 0.473 \\
       
        \textbf{phi3-C} & 57.06 & \textbf{70.93} & \textbf{0.913} &\textbf{0.615} \\
         \hline
    \end{tabular}
   \end{adjustbox}
    \caption{Simplification quality on CEFR-SP-Test per levels; T5-grade, FUDGE and proposed method were evaluated using subsets of specific levels (A, B and C level references, respectively).}
    \label{tab:cefr2}
\end{table}

\begin{table}[t!]
    \small
    \centering
   \begin{adjustbox}{width=1\linewidth}
    \begin{tabular}{l L L L L} \hline
     \textbf{TURK} & \text{LENS}  &  \text{SALSA}  &  \text{Fluency}  &  \text{Adequacy}     \\ \hline
     Reference & 35.20 &  64.96 & 0.732 & 0.901 \\ \hline
       ACCESS  & 49.90 & 62.68 & 0.576 & 0.780 \\
       DMASS   & 46.52 & 58.97 & 0.515 & 0.665 \\
      DRESS  & 59.76 & 62.63& 0.807 & 0.615 \\
      DRESS-LS   &60.56 & 62.92 & 0.838 & 0.657 \\
        EditNTS   &57.71 & 64.86 & 0.752 & 0.710 \\
       IterativEdit   & 37.35 & 49.74 & 0.409 & 0.607 \\
      
     \hline
        phi3-3b-vanilla  & 65.08 & 71.93 & 0.830 & 0.807 \\
        \textbf{phi3-A} & 64.92 & \textbf{73.68} & 0.720 & 0.708 \\
        \textbf{phi3-B} & \textbf{70.25} & 69.05 & 0.855 & \textbf{0.952} \\
        \textbf{phi3-C} & 62.24 & 70.43 & \textbf{0.869} &0.872 \\
         \hline
    \end{tabular}
   \end{adjustbox}
    \caption{Simplification quality on TurkCorpus; all models evaluated on the entire sentences as TurkCorpus does not annotate levels.  }
    \label{tab:turk2}
\end{table}

Table~\ref{tab:human} shows the results; the simplicity score is generally high, close to $1$, across models. 
This is expected as the source sentences were generated to be highly complex. %the model simplifies complicated sentences which are designed to be generated as highly complex. 
The adequacy measurement results are consistent with automatic evaluation; identifying our proposed models as the most adequate. 
Furthermore, the proposed method achieves the best controllability on sentence levels compared to the baselines as indicated by significantly higher ``Level'' and ``Prefer'' scores.

\subsection{Ablation Study}
In this section, we show how each part of the proposed rewards contributes to the final performance. We compare the following models: vanilla phi3 model, reward using only target vocabulary counts, reward using dynamically adjusted vocabulary coverage rates, and reward using both dynamic vocabulary coverage rate and sentence levels (proposed method). 
The frequency and diversity evaluation results for A and B level models are presented in Fig.~\ref{fig:ablation}. Complete results can be found in Appendix \ref{sec:appendix-results}.
It can be seen that changing the simple match count reward to a dynamically adjusted reward indeed encourages the model to increase the entropy inside the target vocabulary and largely improve the vocabulary diversity. 

\begin{table}[t!]
    \small
    \centering
   \begin{adjustbox}{width=1\linewidth}
    \begin{tabular}{l L L L L L} \hline
    Model & \text{Simplicity}  & \text{Adequacy}  & \text{Fluency}  & \text{Prefer} & \text{Level} \\ \hline
     Reference & 1.00 & 0.89 & 0.99 & 0.87 & $--$ \\
     T5+grade-A & 0.83 & 0.16 & 0.47 &0.40 &0.10 \\
     T5+grade-B & 0.90 & 0.13 & 0.50 &0.43  &0.17\\
     T5+grade-C & 0.80 &0.16& 0.60
     &0.40 &0.17 \\
     FUDGE-A & 1.00 & 0.50 & 0.80 & 0.50 &0.43\\
     FUDGE-B & 0.96 & 0.43 & 0.83 &0.57&0.47\\
     FUDGE-C & 1.00 & 0.47 & 0.83&0.57&0.33\\
     \hline
     Ours phi3-A & 1.00 &  0.76 & 0.90 & 0.67 &0.83 \\ 
     Ours phi3-B & 1.00 & 0.83 & 0.90 &0.70 &0.63 \\
     Ours phi3-C & 0.96 & 0.80 & 1.00 &0.80 &0.60 \\ \hline
    \end{tabular}
   \end{adjustbox}
    \caption{Human evaluation results}
    
    % "Level" denotes if the sentence level of the generated simplification, which was guided by the sentence-level reward model during training, matches the level of the gold reference.}
    \label{tab:human}
\end{table}

\begin{figure}[t!]
    \centering
    \includegraphics[width=\linewidth]{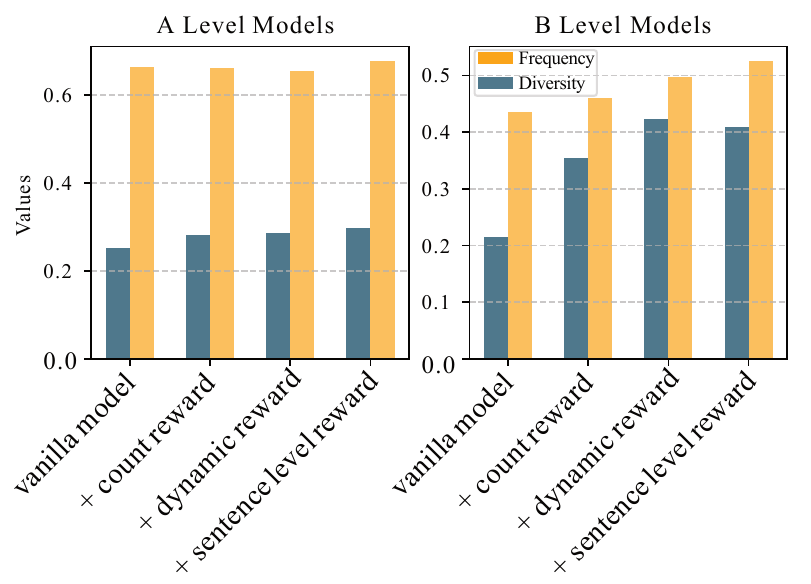}
    \caption{
    Reward effects on target vocabulary coverage
    % Vocabulary targeting w/ different rewards
    }
    \label{fig:ablation}
\end{figure}

\section{Conclusion}
In this paper, we target ESL learners as audiences for text simplification aiming to facilitate the learning process of the foreign language. 
Referring to the input hypothesis and frequency effect theory in L2 learning, we propose a reinforcement learning method on LLM to control the simplification model to generate outputs that satisfy the vocabulary and sentence level constraints. 
Experiment results show that the proposed method can increase the target vocabulary coverage in the generated simplifications, and human evaluation results confirmed that the simplified texts generally preserve the targeted CEFR levels. 

In practice, different individuals have varied levels of knowledge for the language. 
We plan to extend the method to generate individual learner-targeted personalized simplifications in the future.

\section*{Limitations}
This work assumes the target vocabulary for the learner is accessible, which in reality may not be the case as the target vocabulary varies with learner individuals and has to be first estimated. 
Although it is out of the scope of this paper, this direction constitutes our future work. 
Besides, currently, we do not control the frequency to a specific number, such as $95\%$ $i$ level and $5\%$ $i+1$ level, which is an important aspect to consider according to the L2 learning theory. 

The control for target vocabulary and sentence is implemented individually for different levels rather than using one model altogether, causing heavier computational loads. 
In future, we seek to improve the design of reward model to integrate rewards for different proficiency levels into one model, and explore for a finer control over the frequency of the generated vocabulary.

% \bibliography{anthology, latex/custom}

\appendix

\section{Evaluation Metrics}\label{sec:appendix-metrics}
In this section, the evaluation metrics are explained in details: fluency, adequacy, target vocabulary frequency, diversity and target sentence level.

Previous studies have introduced various metrics for evaluating simplicity, which we summarize in Table \ref{tab:metrics}. Among these metrics, SARI \cite{xu-etal-2016-optimizing} is the most commonly employed in the literature. However, recent studies show that SARI may not be an optimal measure for assessing the quality of simplicity \cite{alva-manchego-etal-2021-un, maddela-etal-2023-lens, stodden-etal-2023-deplain}. 
Thus, we chose to use LENS \cite{maddela-etal-2023-lens} and SALSA \cite{heineman-etal-2023-dancing}, two recently proposed metrics, to measure simplicity. 

\begin{table}[h]
    \centering
    \begin{adjustbox}{width=1\linewidth}
    \begin{tabular}{c|c|c} \hline
        Metric & Scope & Reference \\ \hline
        BLEU \cite{papineni-etal-2002-bleu}& semantic similarity & Y \\
        FKGL \cite{kincaid1975derivation} & readability & N \\
        FKBLEU \cite{xu-etal-2016-optimizing} & readability, similarity & Y\\
        SARI \cite{xu-etal-2016-optimizing} & keep, add, delete & Y \\
        D-SARI \cite{sun-etal-2021-document} & keep, add, delete & Y \\
        SAMSA \cite{sulem-etal-2018-semantic}& semantic structural similarity & N \\
        BERTScore \cite{Zhang2020BERTScore} & semantic similarity & Y \\
        SLE \cite{cripwell-etal-2023-simplicity}& human rating + FKGL & N \\
        LENS \cite{maddela-etal-2023-lens} & human rating & Y \\
        SALSA \cite{heineman-etal-2023-dancing}& human rating & N \\ \hline
    \end{tabular}
    \end{adjustbox}
    \caption{Metrics used in recent literature. Scope denotes the aspect that the metric aims to evaluate, and reference indicates whether the metric is computed based on references or not.}
    \label{tab:metrics}
\end{table}

For adequacy and fluency, the ideal approach is human evaluation; however, this is impractical due to the large dataset size. Instead, we employed large language models to assess these two aspects. With a capable language model $f$, the generated simplification sentence $s$ is evaluated as:
\begin{equation}
    score(s) = \sum_{v \in V_y }f(v \mid (pmt, s)
\end{equation}
where $pmt$ is a prompt designed for the model to output "yes" if the model evaluate $s$ to be adequate or fluent, and $V_y$ is a vocabulary subset for "yes" with $ V_y = \{\text{YES}, \text{Yes}, \text{yes}\}$. We use Llama-3-8b-instruct\footnote{https://llama.meta.com/llama3/} model as the evaluation model in our experiment.

To measure target vocabulary frequency, we took the ratio between the total count of matched target words and the total generated words.
% \math
\begin{equation}
      \sum_{j=1}^{m} \sum_{k=1}^{n} \text{count}(C_j,  \text{seq}_k) / \sum_{k=1}^{n}\text{count}(\text{token},  \text{seq}_k)
\end{equation}
To measure vocabulary diversity, we took the ratio between the number of matched words and number of words in the word list.
% \math
\begin{equation}
    \sum_{j \in D} \mathbf{1} \left( \bigvee_{k=1}^{n} \mathbf{1}_{C_j}( \text{seq}_k) \right)/|D|
\end{equation}

\section{Complex Sentence Generations}\label{sec:appendix-complex}
To generate complex sentences, we prompted the GPT-4 model\footnote{https://openai.com/index/gpt-4/} to rephrase sentences of varied levels into highly complex sentences. To ensure the diversity of the generated complex data, we initially created a variety of seed prompts manually and instructed GPT-4 to generate additional prompts based on these seed prompts. GPT-4 was then prompted to generate complex sentences based on these diversified prompts.
The 5 manually written seed prompts and 10 model generated prompts are presented in Table \ref{tab:prompts}. In total 15 prompts were used to generate complex sentences, for each generation, one of the prompts was randomly selected. We present samples of the generated complex sentences together with simplifications in Table \ref{tab:simplification1} and Table \ref{tab:simplification2}.

\section{Training Details}\label{sec:appendix-training}

\paragraph{Implementation Details of Baselines}
We implemented the baseline models using the transformers library\footnote{https://huggingface.co/docs/transformers/en/index}. T5-s2s models require parallel corpus of complex-simple sentences, for which we used the pseudo-parallel sentences of generated complex sentences and their original sentences, and prepended level tokens for level controlling during training and evaluation. We implemented the FUDGE simplification control model with a Llama-3-8b-instruct model as the generation model, and its logits during the inference were adjusted using the CEFR level classification model released by \cite{arase-etal-2022-cefr}.

\paragraph{Implementation Details of Proposed Method}
The PPO training algorithm was implemented using the TRL library\footnote{https://huggingface.co/docs/trl/main/en/index} with a learning rate of $3e-5$. 
For the dynamic reward model used in the training, we set the $\alpha$ to be $1.2$, as we found a value slightly bigger than $1$ was shown to have better performance empirically; the reward for phrases is always set to be $1.5$ times more than words to reward more on the phrase generation.  For the overall reward, $\lambda$ was set to be $1.5$ to compensate for the vocabulary reward penalty, and $\gamma$  was set to 1. During training, we used the following prompt for the model to generate simplifications: ``Given a complex sentence \{\}, generate a simplified version for it:''.

\begin{figure}
    \centering
    \includegraphics[width=\linewidth]{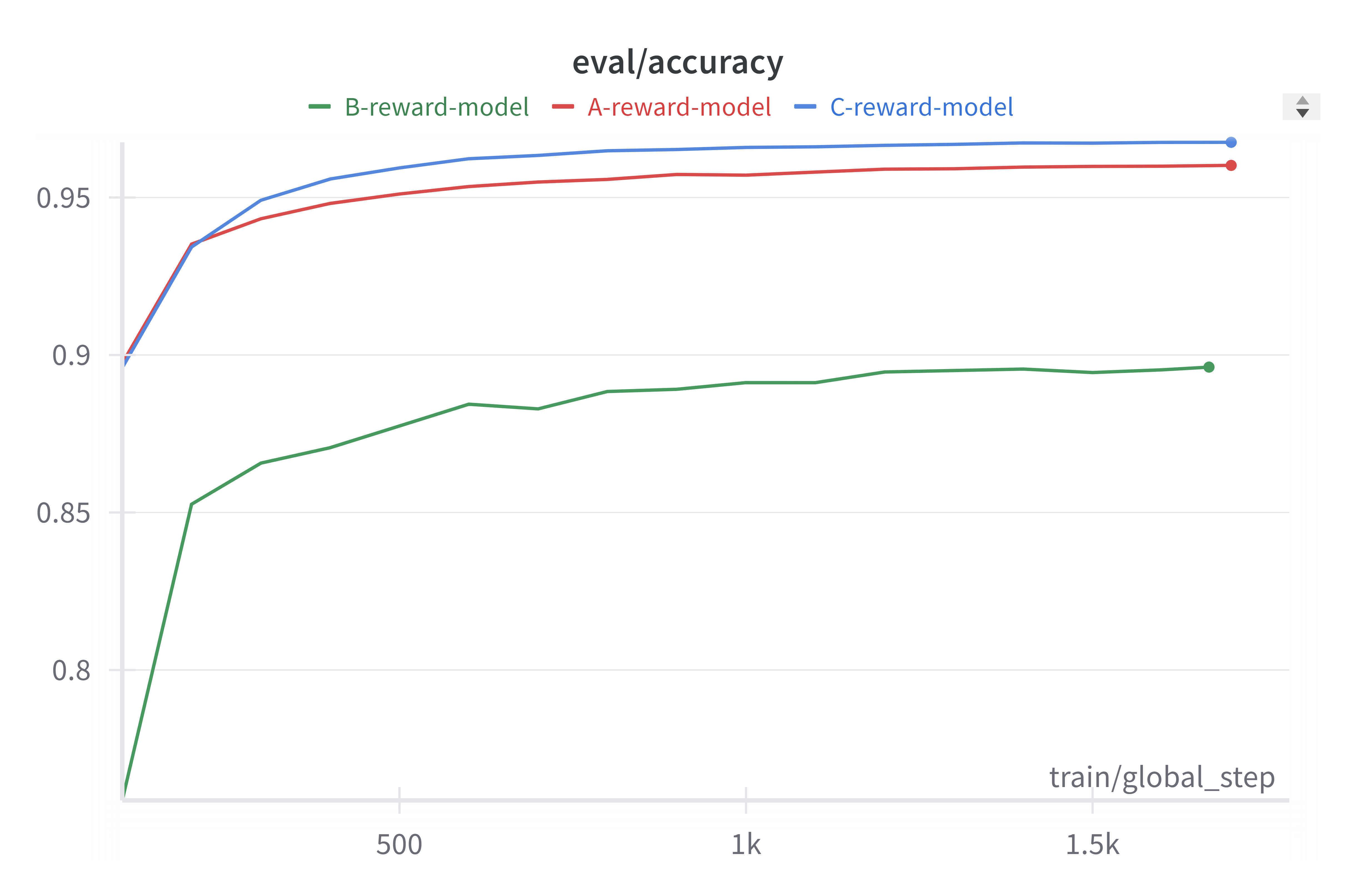}
    \caption{Sentence level reward model evaluation accuracy}
    \label{fig:reward-performance}
\end{figure}

\begin{figure}[h]
    \centering
    \includegraphics[width=\linewidth]{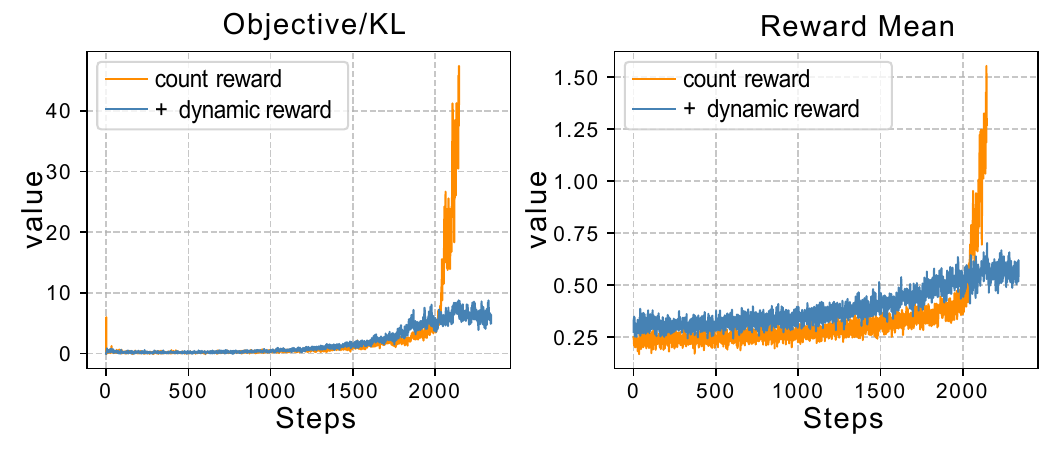}
    % \caption{KL and Reward Change During Training}
    \caption{Training stability w/wo dynamic reward}
    \label{fig:reward}
\end{figure}

\paragraph{Training Performance}
The performance of the sentence level reward model is shown in Fig.~\ref{fig:reward-performance}. 
% Fig. \ref{fig:vocab_all} shows the vocabulary targeting results with different rewards on all three levels of models.
Fig.\ref{fig:reward} shows mean reward and KL change over the training steps with and without the dynamic reward. Objective/KL indicates the deviation of the simplification model from the reference model, and an absurdly high KL indicates model collapse; a burst in mean reward indicates model collapses and only produces a limited set of vocabulary. It can be observed that using the dynamic reward helps stabilize the training, while using the match count alone causes the model to be over-optimized and collapse to a limited vocabulary subset.

\section{Human Evaluation Details}\label{sec:appendix-humaneval}
\begin{figure}[ht]
    \subfigure[Screenshot of annotation guidelines shown to the evaluators]{
    \includegraphics[width=\linewidth]{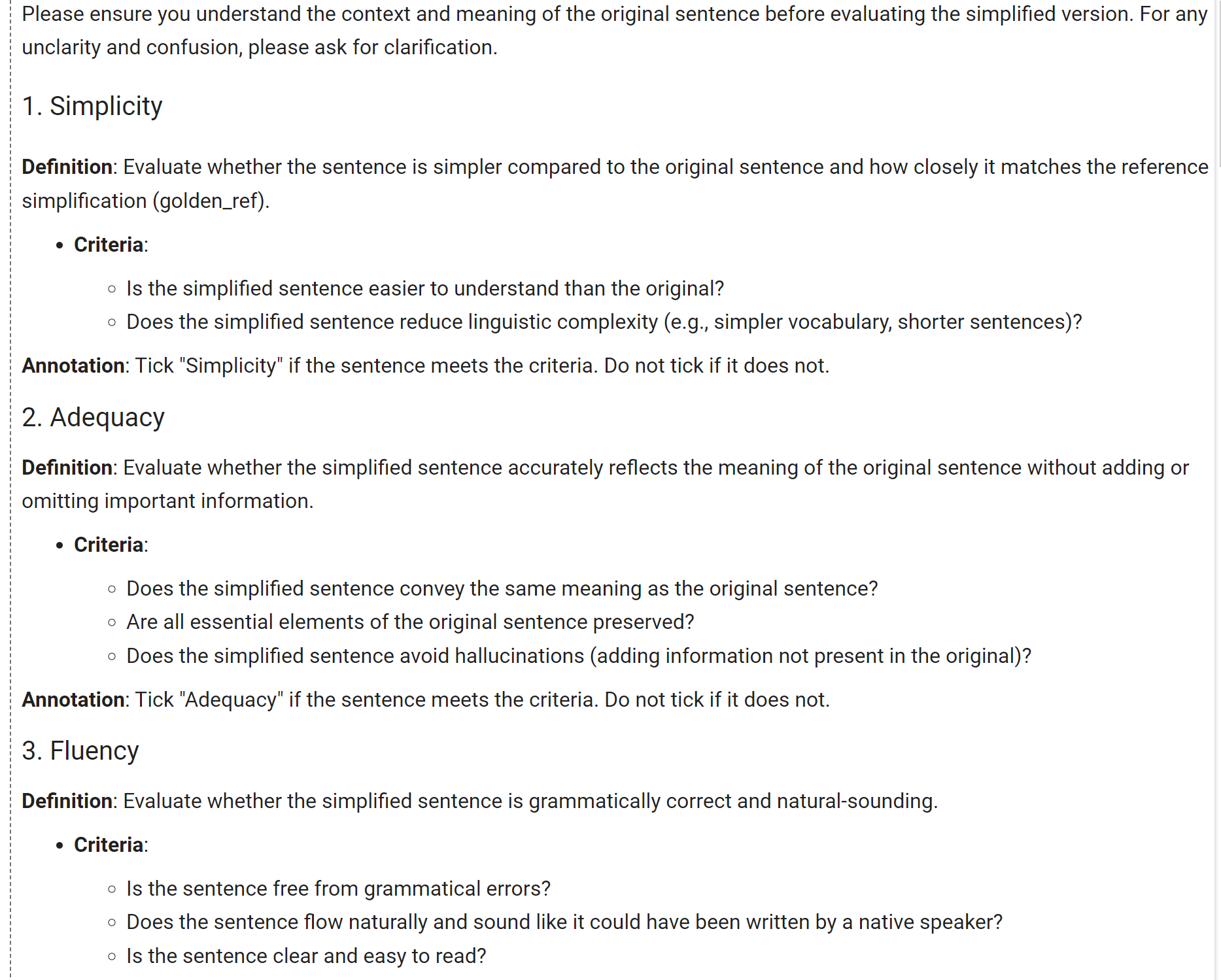} \label{fig:guide1}}

    \subfigure[Screenshot of annotation interface shown to the evaluators]{
    \includegraphics[width=\linewidth]{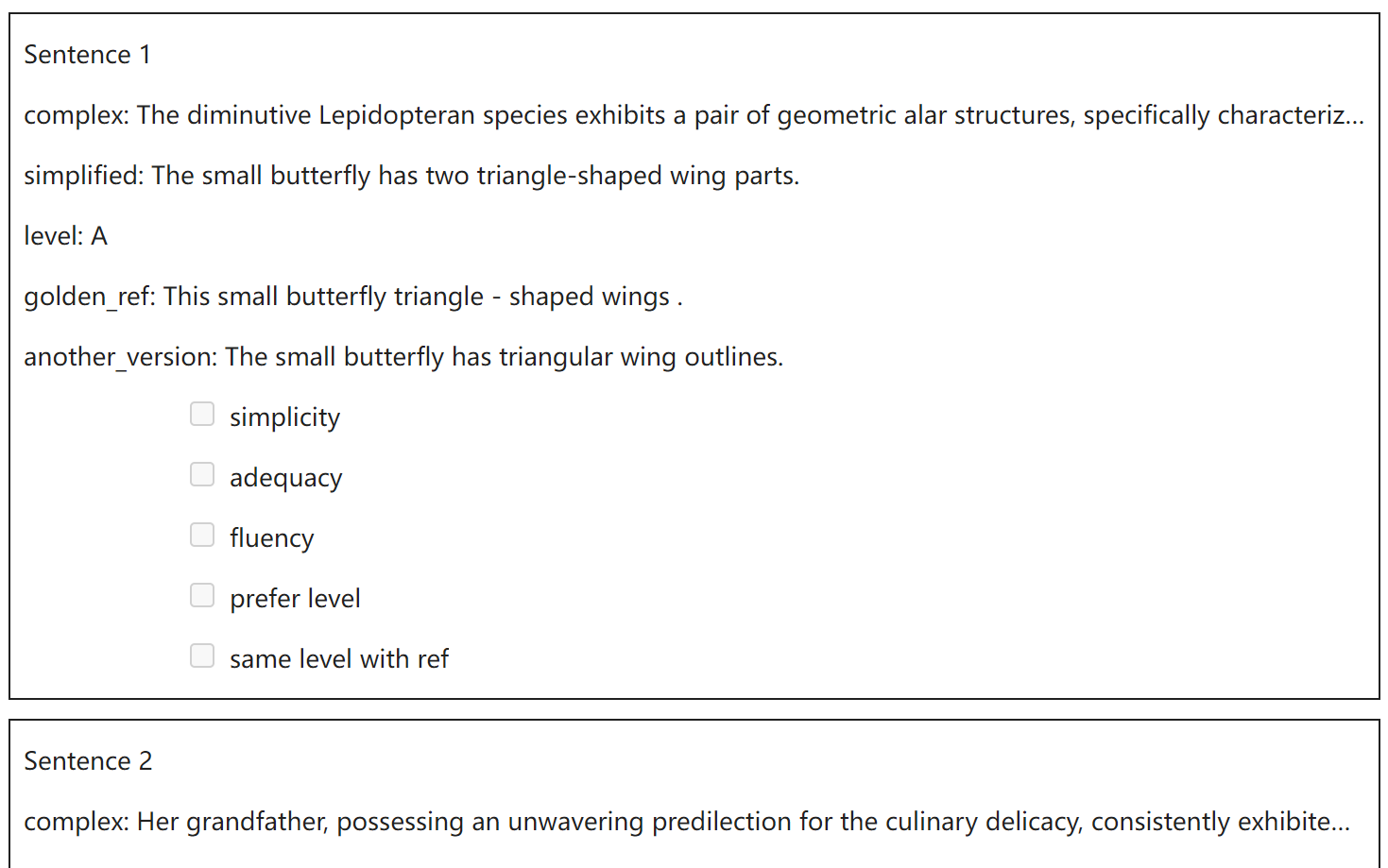}
    \label{fig:interface}}
    \caption{Screenshot of annotation guidelines}
\end{figure}

In this section, the annotation guidelines that evaluators used to evaluate the generated simplifications as well as the annotation interface are presented. 
The annotation guidelines contain the definition of the aspect to be evaluated, the criteria for the evaluation and indications for the annotation interface. The annotation interface is designed to be a binary-choice form, for each aspect to be evaluated, the evaluators chose to tick to indicate that the simplification contains the aspect to be evaluated, and does not satisfy the aspect otherwise. The evaluation results are then used to calculate the binary accuracy of the aspects to be evaluated. 
The evaluation guidelines are shown in Fig. \ref{fig:guide1} and the evaluation interface is shown in Fig. \ref{fig:interface}.

\section{More Evaluation Results}\label{sec:appendix-results}
\begin{figure}[h]
    \centering
    \includegraphics[width=\linewidth]{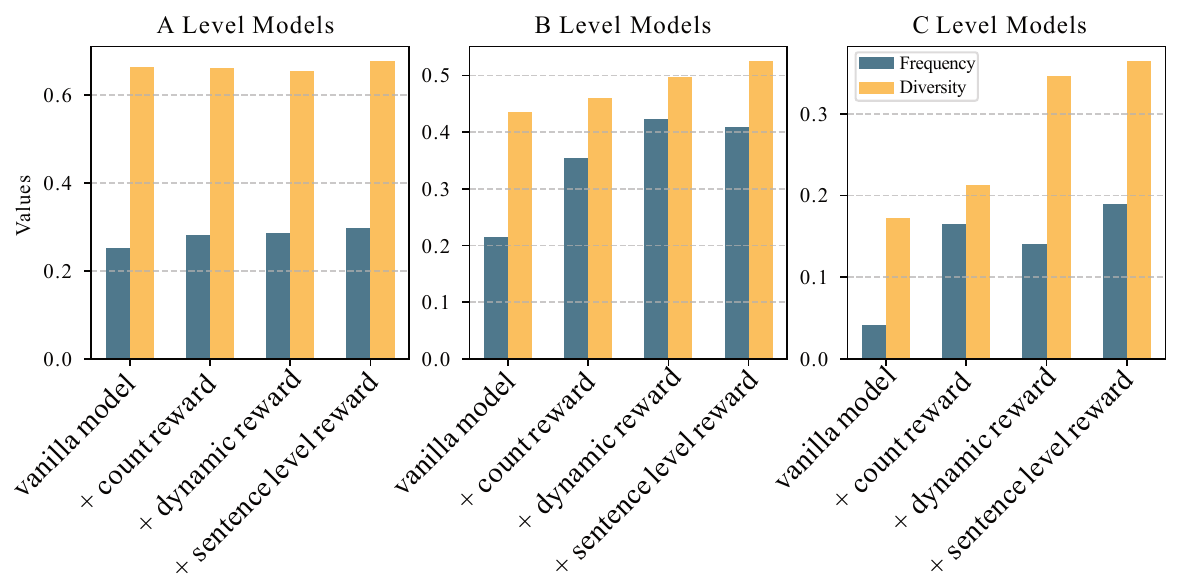}
    \caption{Vocabulary targeting w/ different rewards, all levels}
    \label{fig:vocab_all}
\end{figure}

Fig. \ref{fig:vocab_all} shows the ablation study results on all three levels of models.

Tables \ref{tab:simplification1} and \ref{tab:simplification2} present example outputs: the complex sentences, reference sentences, and simplified sentences. 
For each complex sentences, there are three versions of the simplified sentences, corresponding to A, B and C levels generated by different models targeting the corresponding level, respectively. 
% We also show the original labeled sentence level for the reference sentence.

\begin{table*}[]
    \centering
    \begin{tabular}{{|p{0.9\textwidth}|}} \hline 
     Manual Composed Prompts  \\ \hline
     You are an expert in academic writing, renowned for your ability to compose intricate and sophisticated sentences. Please rephrase the following sentence,so that it's a complex, hard to follow sentence that would usually appear in a journal article, without loss of original meaning: {} \\\hline
     You are an experienced English teacher. Please rephrase the following sentence,to make it a complicated, very hard sentence to read that a English learner may encounter in daily reading, without loss of original meaning: {}\\\hline
    You are a successful postmodernism theater and book critic. You used varied writing styles in your articles. Please rephrase the following sentence,to make it a complex and very difficult to understand sentence,without loss of original meaning: {}\\\hline
    You are a philosopher and literature professor. You usually make intricate perception and sharp insight in your writing.  Please rephrase the following sentence,to make it a short but complex and very hard to follow,without loss of original meaning: {}\\\hline
    You are an editor of social and financial news and journals. Please rephrase the following sentence,so that the sentence has complex compositions and advanced words, that normal readers cannot understand, without loss of original meaning: {}\\ \hline
    Model Generated Prompts  \\ \hline
    You are a legal scholar with extensive experience in drafting complex legal documents. Please rephrase the following sentence,to make it a complex and legally intricate sentence,without loss of original meaning: {}\\\hline
    You are a renowned scientist known for writing dense and comprehensive research papers. Please rephrase the following sentence,to make it a complex and highly technical sentence,without loss of original meaning: {}\\\hline
    You are a seasoned journalist known for crafting elaborate and detailed investigative reports. Please rephrase the following sentence,to make it a complex and deeply investigative sentence,without loss of original meaning: {}\\\hline
    You are a literary critic who writes for a prestigious literary journal, known for your sophisticated language. Please rephrase the following sentence,to make it a complex and highly sophisticated sentence,without loss of original meaning: {}\\\hline
    You are a historian known for your detailed and intricate historical analyses. Please rephrase the following sentence,to make it a complex and historically detailed sentence,without loss of original meaning: {}\\\hline
    You are an expert in technical writing, specializing in creating elaborate and detailed user manuals. Please rephrase the following sentence,to make it a complex and technically detailed sentence,without loss of original meaning: {}\\\hline
    You are a linguist with expertise in creating intricate and multifaceted linguistic analyses. Please rephrase the following sentence,to make it a complex and linguistically intricate sentence,without loss of original meaning: {}\\\hline
    You are a political theorist known for your dense and intricate political analyses. Please rephrase the following sentence,to make it a complex and politically intricate sentence,without loss of original meaning: {}\\\hline
    You are an economist renowned for your detailed and complex economic analyses. Please rephrase the following sentence,to make it a complex and economically detailed sentence,without loss of original meaning: {}\\\hline
    You are a theologian known for your intricate and deeply philosophical theological writings. Please rephrase the following sentence,to make it a complex and theologically intricate sentence,without loss of original meaning: {}\\\hline

\end{tabular}
    \caption{Prompts used to generate complex sentences}
    \label{tab:prompts}
\end{table*}

\begin{table*}[]
    \centering
    \begin{tabular}{|p{0.9\textwidth}|}\hline
       
        Complex Sentence \\
        Let us endeavor to delve into the intricacies and nuances of the text, striving to comprehend the underlying themes and implications inherent within, as we embark on this journey of intellectual exploration.\\\hline
        Reference, level: A \\   
        Let 's try to read .\\\hline
        Simplifications  \\
        Let's try to understand the text well. \\
        Let's try to understand the text's main ideas and meanings as we read.\\
        Let's understand the text's themes and meanings.\\ \hline
        Complex Sentence \\
        Are you experiencing any concerns or apprehensions regarding the composition, quality, or safety of the food you consume?\\\hline
        Reference, level: A \\   
        Do you worry about what you eat ?\\\hline
        Simplifications  \\
        Do you have worries about the food you eat? \\
        Do you have worries about the food you eat's quality or safety? \\
        Do you have concerns about your food's composition, quality, or safety?\\\hline
    %    Complex Sentence \\
    %     In a commendable display of relentless dedication and tireless effort, my husband engages in his professional endeavors with an exceptional degree of industriousness, thereby exemplifying a work ethic that is both remarkable and inspiring.\\\hline
    %     Reference, level: A \\   
    %     My husband works hard .\\\hline
    %     Simplifications  \\
    %     My husband works very hard at his job.\\
    %     My husband works hard and does a great job at his job.\\
    %     My husband consistently shows outstanding work dedication.\\\hline
    %     Complex Sentence \\
    % Within the confines of the structural barriers, an incessant cacophony of scraping sounds revealed the furtive activities of diminutive rodents, whose movements left an indelible mark upon the silent fabric of the edifice.\\\hline
    % Reference, level: A \\   
    % There were mice scratching in the walls .\\\hline
    % Simplifications  \\
    % In a building, small rodents make scratching noises and leave marks.\\
    % Inside the building, small rodents made scratching noises and left marks.\\
    % Structural barriers echoed with rodent noises, leaving noticeable trails.\\\hline
    Complex Sentence \\
The considerable distance, compounded by Jamie's current condition of pregnancy, which inexorably engenders a state of increased fatigue, renders the prospect of ambulation to said location prohibitively challenging for her.\\\hline
Reference, level: B \\   
It is too far for Jamie to walk to , especially because she is pregnant and easily exhausted .\\\hline
Simplifications  \\
Jamie is too tired to walk far because she is pregnant.\\

Jamie's pregnancy makes it very hard for her to walk to the location due to the long distance.\\
Jamie's pregnancy leads to fatigue, making it hard for her to walk to the distant place.\\\hline

Complex Sentence \\
With meticulous care, the young girl positioned the doll on her bed, her actions imbued with a quiet reverence, before softly pressing her lips to its cheek, an act infused with such tenderness that one might easily imagine she believed the doll to be a living boy.\\\hline
Reference, level: B \\   
The little girl placed the doll carefully on her bed and kissed its cheek as gently as if it were a real boy .\\\hline
Simplifications  \\
The girl put the doll on her bed and kissed it gently.\\
The girl carefully placed the doll on her bed, kissed its cheek gently, and seemed to love it like a boy.\\
The girl gently kissed the doll out of affection.\\\hline

    \end{tabular}
    \caption{Complex sentences, references and model simplified sentences in A, B, C levels, respectively}
    \label{tab:simplification1}
\end{table*}

\begin{table*}[]
    \centering
    \begin{tabular}{|p{0.9\textwidth}|}\hline

        Complex Sentence \\
        Numerous physical characteristics of alkenes and alkanes exhibit remarkable congruity; manifesting as chromatically imperceptible, possessing nonpolarity, and exhibiting combustibility.\\\hline
        Reference, level: C \\   
        Many of the physical properties of alkenes and alkanes are similar : they are colorless , nonpolar , and combustable .\\\hline
        Simplifications  \\
        Alkenes and alkanes have similar physical traits. They are hard to see color differences, not attracted to water, and burn easily.\\
        Alkenes and alkanes have similar physical traits. They are colorless, nonpolar, and burn easily.\\
        Alkenes and alkanes share traits like colorlessness, nonpolarity, and flammability.\\\hline

        By the month of May in the year 2008, the phenomenon known as Rickrolling had ascended to the status of an internet sensation, infiltrating the digital landscape with such pervasive enthusiasm that it swiftly garnered the attention of mainstream media outlets, which began to extensively cover this viral trend.\\\hline
    Reference, level: C \\   
    By May 2008 , Rickrolling had become an internet hit or phenomenon and very soon rickrolling was being covered on mainstream media .\\\hline
    Simplifications  \\
    In 2008, Rickrolling became very popular on the internet and in the news.\\
    In 2008, Rickrolling became a big internet trend.\\
    In 2008, Rickrolling became an internet sensation, leading to widespread media coverage.\\\hline

    \end{tabular}
    \caption{Complex sentences, references and model simplified sentences in A, B, C levels, respectively}
    \label{tab:simplification2}
\end{table*}

\end{document}